\documentclass{ieeeaccess}
\usepackage{cite}
\usepackage{amsmath,amssymb,amsfonts}
\usepackage{algorithmic}
\usepackage{graphicx,subfig,float}
\usepackage{textcomp}
\usepackage{dirtytalk}
\usepackage{caption}
\usepackage{hyperref}
\usepackage{booktabs}

\newcommand{\D}{\mathbb{D}}
\newcommand{\x}[1]{x^{(#1)}}

\newcommand{\E}[1]{\mathbb{E}_{#1}}

\DeclareMathOperator{\EX}{\mathbb{E}}

\def\BibTeX{{\rm B\kern-.05em{\sc i\kern-.025em b}\kern-.08em
    T\kern-.1667em\lower.7ex\hbox{E}\kern-.125emX}}
\begin{document}
\history{Date of publication xxxx 00, 0000, date of current version xxxx 00, 0000.}
\doi{10.1109/ACCESS.2017.DOI}

\title{Enhancing variational generation through self-decomposition}
\author{\uppercase{Andrea Asperti}\authorrefmark{1}, 
\uppercase{Laura Bugo}\authorrefmark{1}, and \uppercase{Daniele Filippini}\authorrefmark{1}}
\address[1]{University of Bologna, Department of Informatics: Science and Engineering (DISI)}

\markboth
{Author \headeretal: Preparation of Papers for IEEE TRANSACTIONS and JOURNALS}
{Author \headeretal: Preparation of Papers for IEEE TRANSACTIONS and JOURNALS}

\corresp{Corresponding author: Andrea Asperti (e-mail: andrea.asperti@unibo.it).}

\begin{abstract}
In this article we introduce the notion of Split Variational Autoencoder (SVAE), whose output $\hat{x}$ is obtained as a weighted sum $\sigma \odot \hat{x_1} + (1-\sigma) \odot \hat{x_2}$ of two generated images $\hat{x_1},\hat{x_2}$, and
$\sigma$ is a {\em learned} compositional map. The composing images $\hat{x_1},\hat{x_2}$, as well as the 
$\sigma$-map are automatically synthesized by the model. The network is trained as a usual Variational Autoencoder with a negative loglikelihood loss 
between training and reconstructed images. No additional loss is required
for $\hat{x_1},\hat{x_2}$ or $\sigma$, neither any form of human tuning. The decomposition is
nondeterministic, but follows two main schemes, 
that we may roughly categorize as either \say{syntactic} or \say{semantic}. 
In the first case, the map tends to exploit the strong correlation between adjacent pixels, splitting the image in
two complementary high frequency sub-images. 
In the second case, the map typically
focuses on the contours of objects, splitting the image in
interesting variations of its content, with more marked and
distinctive features. 
In this case, according to empirical observations, the Fr\'echet Inception Distance (FID) of 
$\hat{x_1}$ and $\hat{x_2}$ is usually lower (hence better) 
than that of $\hat{x}$, that clearly suffers from being the average of the former. In a sense, a SVAE forces the Variational Autoencoder to {\em make choices}, in contrast with its intrinsic tendency to {\em average} between alternatives with the aim to minimize the reconstruction loss towards a specific sample.
According to the FID metric, our technique, 
tested on typical datasets such as Mnist, Cifar10 and CelebA, allows us to outperform all previous purely variational architectures (not relying on normalization flows). 
\end{abstract}


\titlepgskip=-15pt

\maketitle

\section{Introduction}
Generative modeling (see e.g. \cite{generative_introduction} for an introduction) is one of the most fascinating problems in Artificial Intelligence, with many relevant applications in different areas comprising computer vision, natural language processing, medicine or reinforcement learning. The goal is not only to be able to sample new realistic examples starting from a given set of data, but to gain insight in the data manifold, and the way a neural network is able to extract and exploit the characteristic features of data. In case of high-dimensional data, the generative problem can only be addressed by means of Deep Neural Networks, and this topic led to tremendous research in many different directions, particularly nourishing the recent field of unsupervised representation learning. 

Among the different kind of generative models which have been investigated, Variational Autoencoders  (VAEs) \cite{Kingma13,RezendeMW14} have always exerted a particular fascination \cite{VAEbiomed,variationsVAE,astrovader}, mostly due to their strong theoretical foundations, that will be briefly recalled 
in Section~\ref{sec:background}. Unfortunately, results remained below expectations, and the generative quality of VAEs is systematically outperformed by different generative techniques like e.g. Generative Adversarial Networks. 

A particularly annoying problem is that VAEs produce images with a characteristic blurriness, very hard to be removed with traditional techniques \cite{blind_deblurring,deblurring1}.
The source of the problem is not easy to identify, but it is likely 
due to {\em averaging}, implicitly underlying the VAE frameworks and, more generally, any autoencoder approach. As observed in \cite{tutorial-GAN}. In presence of multimodal output, a loglikelihood objective typically results in averaging and hence blurriness.
A GAN does not have this problem, since its goal is to fool the Discriminator, not to reconstruct a given input. 
Since Variational Autoencoders are intrinsically multimodal, both 
due to dimensionality reduction, and to the sampling process during
training, a certain amount of blurriness is unfortunately expected.

\begin{figure*}[ht]
\begin{center}
\includegraphics[width=\textwidth]{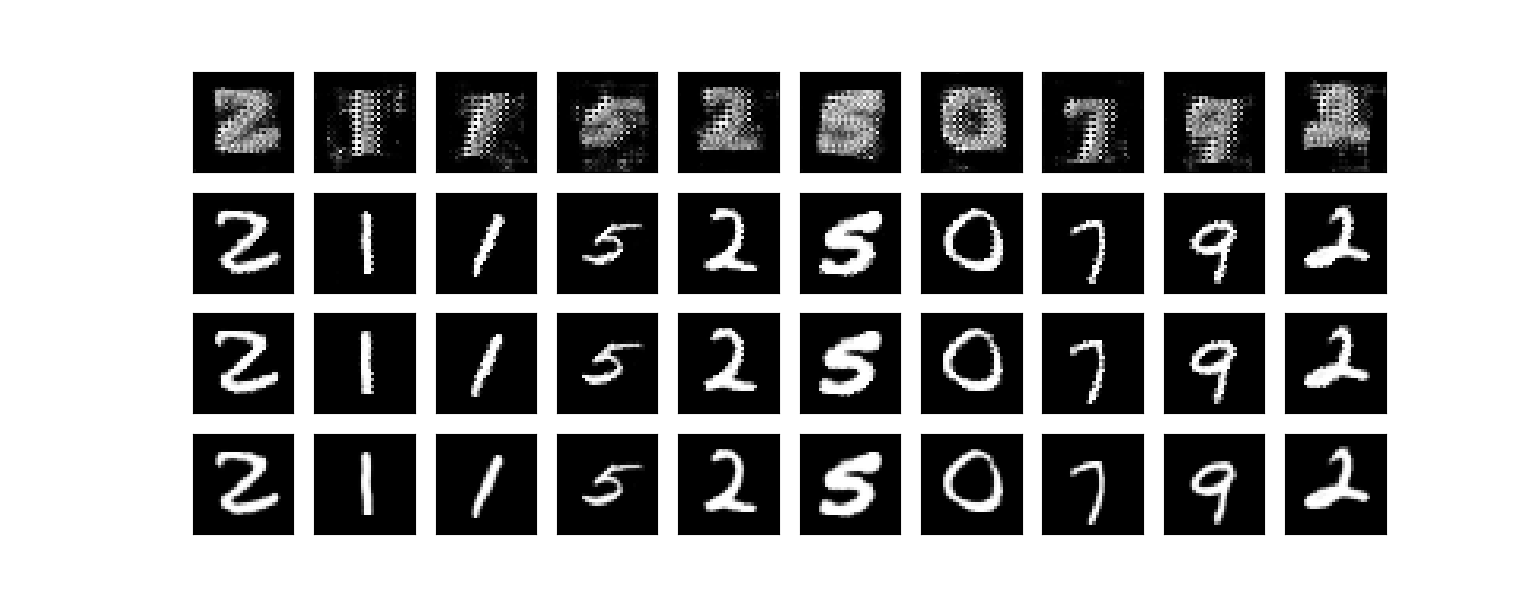}
\caption{Example of ``syntactic'' decomposition for Mnist. In the first line we
have the $\sigma$ map, then, in order, $\hat{x_1}$, $\hat{x_2}$, and finally 
$\hat{x} = \sigma\odot \hat{x_1} + (1-\sigma) \odot \hat{x_2}$ (similarly for
the other analogous pictures).
In this case, the image is decomposed in complementary subimages at high-frequency. This usually helps to decorrelate adjacent pixels in the latent encoding.
\label{fig:mnist_syntactic}}
\end{center}
\end{figure*}

\begin{figure*}[ht]
\begin{center}
\includegraphics[width=\textwidth]{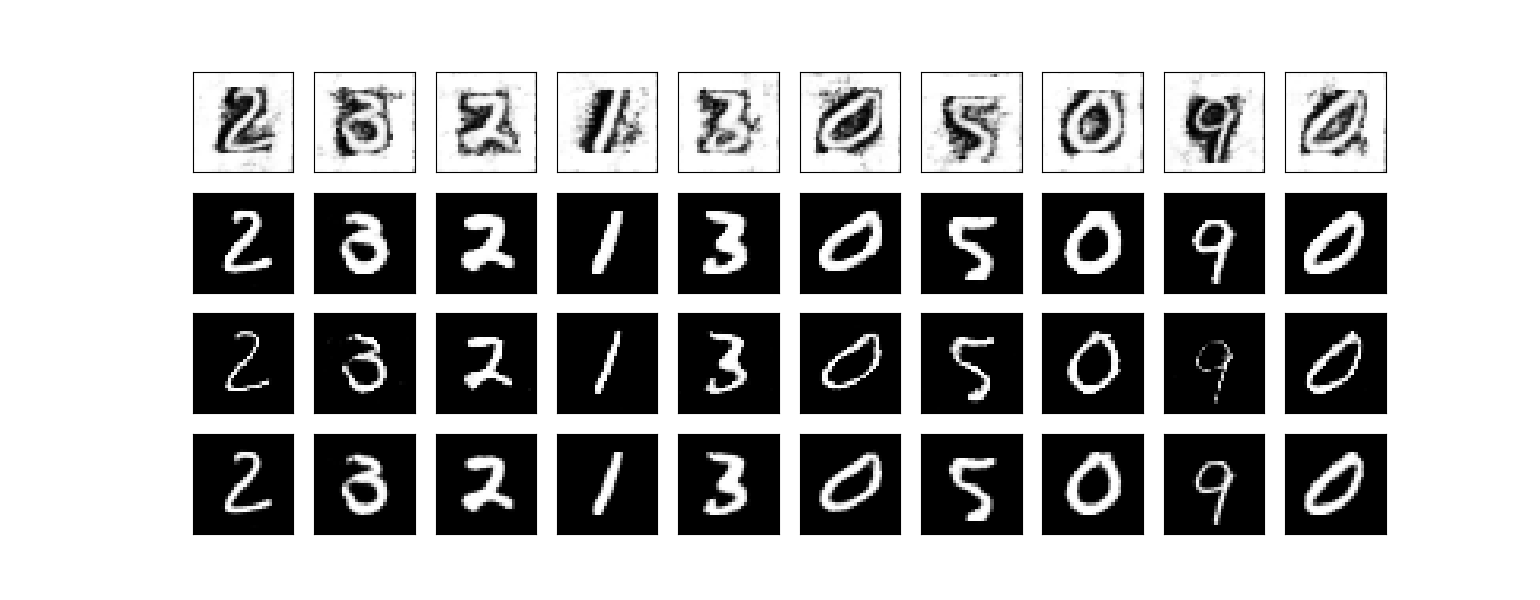}
\caption{Example of ``semantic'' decomposition for Mnist. Digits are usually decomposed in a ``fat'' and a ``thin'' version following the contours of objects. The compound image
$\hat{x} = \sigma \odot \hat{x_1} + (1-\sigma) \odot \hat{x_2}$ is particularly neat and mushy.
\label{fig:mnist_semantic}}
\end{center}
\end{figure*}

Starting from the averaging assumption \cite{varianceloss}, it is natural to try to address
blurriness by offering to the Variational Autoencoder the possibility to create multiple images, and then synthesize a result as a (learned) weighted combination of them. This is precisely what our Split Variational Autoencoder (SVAE) is supposed to do: the generator returns
two images $\hat{x_1}, \hat{x_2}$ and a probability map $\sigma$ with the same spatial dimension of the images, and synthesize a resulting image
$\hat{x} = \sigma \odot \hat{x_1} + (1-\sigma) \odot \hat{x_2}$ where $\odot$ is point-wise multiplication (broadcasted over channels). 
The Autoencoder is trained by minimizing the reconstruction loss between
$x$ and $\hat{x}$, together with the traditional regularization 
component over latent variables. No additional loss is imposed over $\hat{x_1}$, $\hat{x_2}$ or $\sigma$.

\begin{figure*}[ht]
\begin{center}
\includegraphics[width=\textwidth]{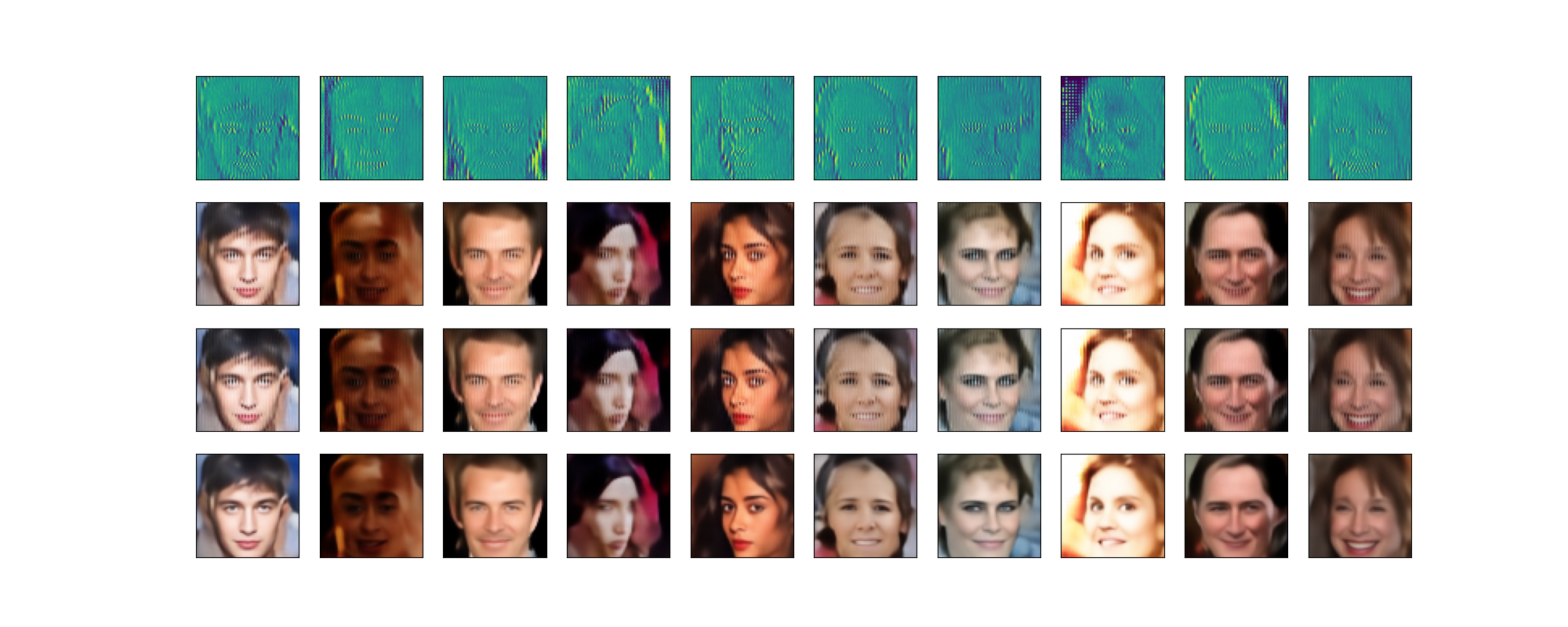}
\caption{Example of ``syntactic'' decomposition for CelebA.
In this case, the FID score for $\hat{x_1}$ and $\hat{x_2}$
is usually bad. Still, the decomposition helps to get a stable 
and robust training, typically resulting in good generative results for
the compound image $\hat{x} = \sigma \odot \hat{x_1} + (1-\sigma) \odot \hat{x_2}$. 
\label{fig:celeba_syntactic}}
\end{center}
\end{figure*}

\begin{figure*}[ht!]
\begin{center}
\includegraphics[width=\textwidth]{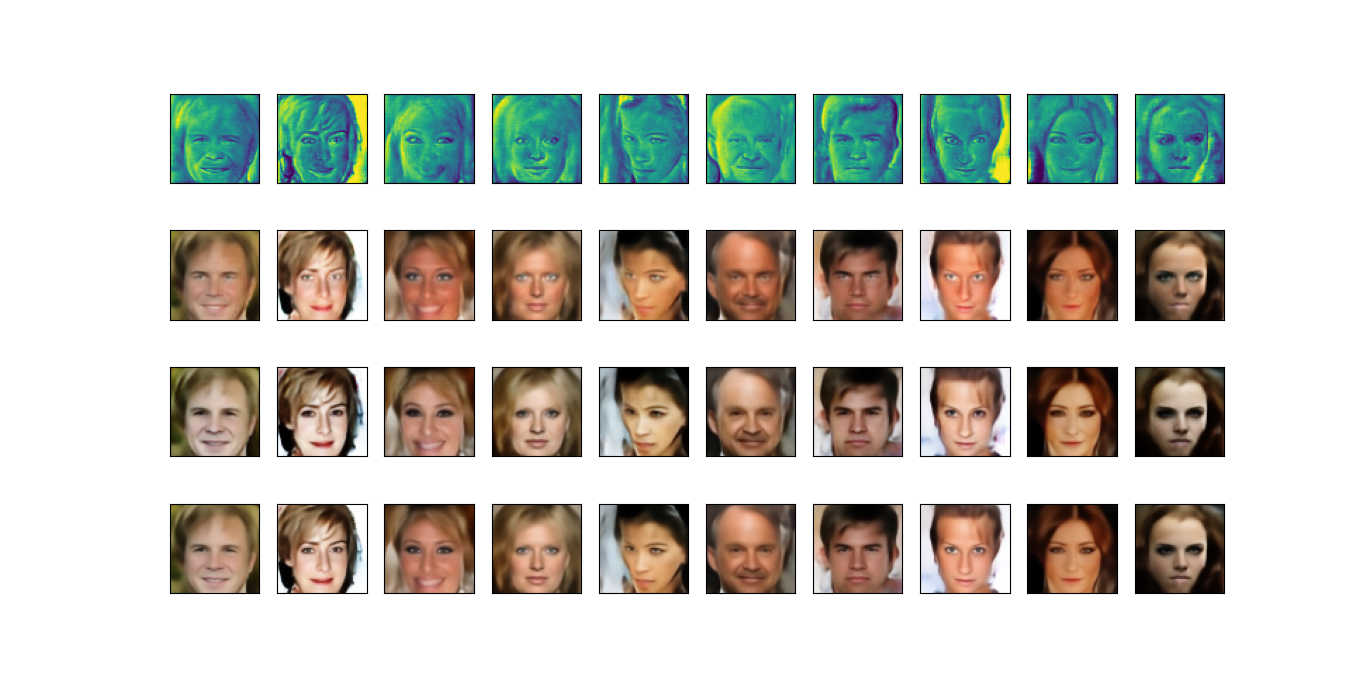}
\caption{Example of ``semantic'' decomposition for CelebA.
This is the most interesting case. The map focuses on contours
of objects, emphasizing them in opposite directions.  
This is frequently rewarding in terms of FID score for $\hat{x_1}$ and $\hat{x_2}$, that are usually better than that of $\hat{x}$.
\label{fig:celeba_semantic}}
\end{center}
\end{figure*}

The resulting decomposition is non deterministic, mostly depending on 
the network architecture and the dimension of the latent space. However, it seems to follow two main schemes, that we call ``syntactic'' (see Figures \ref{fig:mnist_syntactic}, \ref{fig:celeba_syntactic}), and ``semantic'' (see Figures \ref{fig:mnist_semantic}, \ref{fig:celeba_semantic}). In this Figures, the top line is the $\sigma$ map, the second line is $\hat{x_1}$, the third line is $\hat{x_2}$, and in the last line we 
have $\hat{x} = \sigma\odot \hat{x_1} + (1-\sigma) \odot \hat{x_2}$.
Let us also remark that all images in the pictures have been
{\em generated}, not reconstructed. 

In the first case, the map takes advantage of the strong correlation between adjacent pixels splitting the image in
two complementary high frequency sub-images. Each
image has more freedom in filling the ignored parts, easing the generative task.

In the second, even more interesting case,  the map focuses on the {\em contours} of objects, splitting the image in
interesting variations around them, typically resulting in more marked and
distinctive features. 
In this case, the Fr\'echet Inception Distance (FID) of $\hat{x_1}$ and $\hat{x_2}$ may also be lower (hence better) 
than that of $\hat{x}$, that apparently suffers from being the average of the former.

An interesting aspect of SVAEs, and possibly one of reasons behind their effectiveness, is that they allow to work with a number of latent variables sensibly higher than usual, hence implicitly addressing the variable collapse phenomenon \cite{BurdaGS15,overpruning17,Trippe18,PosteriorCollapse,sparsity}. This seems to be an indication that self-splitting is indeed a convenient way to induce the model to synthesize a large number of uncorrelated latent features.


We tested SVAE on typical datasets such as Mnist, Cifar10 and CelebA, and in all cases we observed substantial improvements w.r.t. the ``vanilla'' approach. Excluding models that make use of sophisticated techniques like normalizing flows \cite{glow,NVAE,morrow2020variational} (typically requiring thousands of latent variables, and practically hindering a fruitful exploration of the latent space), SVAE 
outperforms all previous variational architectures.

The code relative to this work can be accessed on Github in the following public repository: \href{https://github.com/asperti/Split-VAE}{\url{https://github.com/asperti/Split-VAE}}
Pretrained weights for the models discussed in the article
are available at the following page: \href{https://www.cs.unibo.it/~asperti/SVAE.html}{\url{https://www.cs.unibo.it/~asperti/SVAE.html}}.

\subsection{Structure of the article}\label{sec:structure}
In Section~\ref{sec:background} we briefly recall the theory behind 
Variational Autoencoder, show their encoder-decoder architecture
(Section~\ref{sec:vanillaVAE}), and discuss some aspects related to the 
dimension of the latent space and the so called variable-collapse phenomenon
(Section~\ref{sec:sparsity}).
Section~\ref{sec:SVAE} introduces the notion of Split Variational Autoencoder
and provides a detailed description of the ResNet-like architecture used in our experiments.
In Section~\ref{sec:setting} we outline our experimental setting, 
discussing the metrics and datasets used for the benchmarks. 
Quantitative results are given in Section~\ref{sec:tables} along with a critical discussion. Ablation investigations are debated in Section~\ref{sec:ablation}. In the Conclusions~\ref{sec:conclusions}, we summarize 
the content of the article and outline research directions for future developments.

\section{Background}
\label{sec:background}
There exist in the literature several good introductions to Variational Autoencoders (VAEs) \cite{tutorial-VAE,VAEKingma,VAEGreen}, so in this section we provide a quite short introduction to the topic, mostly with the purpose to fix notation and terminology.

In a latent variable approach, the probability distribution 
$p(x)$ of a data point $x$ is expressed through marginalization
over a vector $z$ of {\em latent variables}:
\begin{align}
    p(x) = \int_{z} p(x|z)p(z) dz = \E{p(z)} [p(x|z)]
\end{align}
where $z$ is the latent encoding of $x$ distributed with a known distribution $p(z)$ named {\em prior distribution}. 
If we can learn a good approximation of $p(x|z)$ from the data, we can use it to generate new samples via ancestral sampling:

\begin{enumerate}
    \item[\textbullet] sample $z \sim p(z)$.
    \item[\textbullet] generate $x \sim p(x|z)$.
\end{enumerate}

Supposing to have a parametric family of probability distributions $\{ p_\theta(x|z)\}$ (e.g. modelled by a neural network), the goal is to find $\theta^*$ that optimize the 
loglikelihood over all $x \in \D$ (MLE):
\begin{equation}
\begin{array}{rl}
    \theta^* & = \arg\max_\theta \E{\D}[\log p_\theta(x)]\\
     & = \arg\max_\theta \E{\D} \Bigl[ \log \int_{z} p_\theta(x|z) p(z) dz \Bigr]
    \end{array}\label{MLE}
\end{equation}
Addressing directly the previous optimization problem is usually computationally infeasible. For this reason, VAEs exploit another probability distribution $q_\phi(z|x)$ named {\em inference} (or encoder) distribution, expressing the relation between a data point $x$ and its associated latent representation $z$. Hopefully, $q_\phi(z|x)$ should approximate $p_\theta(z|x)$, so that their  Kullback-Leibler divegence
\[D_{KL}(q_\phi(z|x) || p_\theta(z|x)) = \E{q_\phi(z|x)}[\log q_\phi(z|x) - \log p_\theta(z|x) ]\] 
should be small.
Further expanding the previous equation, we get:
\begin{align}
    \nonumber &D_{KL}(q_\phi(z|x) || p_\theta(z|x)) \\
    \nonumber &=  \E{q_\phi(z|x)}[\log q_\phi(z|x)\! - \! \log p_\theta(z|x) ] \\  
   & =
    \nonumber \E{q_\phi(z|x)}[\log q_\phi(z|x)\! -\! \log p_\theta(x|z) \!-\! \log p_\theta(z)\! +\! \log p_\theta(x) ] \\ 
    \nonumber &= D_{KL}(q_\phi(z|x) || p(z))\! -\! \E{q_{\phi}(z|x)} [ \log p_\theta(x|z) ]\! +\! \log p_\theta(x)
\end{align}
Hence,
\begin{equation}
 \begin{array}{c}
  \E{q_{\phi}(z|x)} [ \log p_\theta(x|z) ] -  D_{KL}(q_\phi(z|x) || p(z))\\
  = \log p_\theta(x) - D_{KL}(q_\phi(z|x) || p_\theta(z|x))
  \end{array}
\end{equation}
Recalling that $D_{KL}$ is always positive, we get
\begin{align}
  \nonumber \underbrace{\E{q_{\phi}(z|x)} [ \log p_\theta(x|z) ] -  D_{KL}(q_\phi(z|x) || p(z))}_{\text{ELBO}}  \leq \log p_\theta(x)
\end{align}
stating that the left hand side is a lower bound for the loglikelihood of $p_\theta(x)$, known as 
Evidence Lower Bound (ELBO). 

Since ELBO is more tractable than MLE, it is used as the cost function for training of neural networks. Optimizing ELBO we are jointly improving the loglikelihood of $p_\theta(x)$, and implicitly minimizing the distance between $q_\phi(z|x)$ and $p_\theta(z|x)$.

The ELBO has a form similar to an autoencoder: the inference distribution $q_\phi(z|x)$ encodes the input $x$ to its latent representation $z$, and $p_\theta(x|z)$ decodes $z$ back to $x$. 

For generative sampling, we just exploit
the decoder, sampling latent variables according to the prior 
distribution $p(z)$ (that must be known).

\subsection{Vanilla VAE and its training} \label{sec:vanillaVAE}
In the vanilla VAE, we assume $q_\phi(z|x)$ to be 
a Gaussian (spherical) distribution $G(\mu_\phi(x),\sigma^2_\phi(x))$, 
so that learning $q_\phi(z|x)$ amounts to learning its two first
moments. It is important to know the variance $\sigma^2_\phi(x)$
since during training we need to sample according to $q_\phi(z|x)$.

Similarly, we also assume $p_\theta(x|z)$ to have Gaussian distribution centered around a decoder function $\mu_\theta(z)$.
The functions $\mu_\phi(x)$, $\sigma^2_\phi(x)$ and $\mu_\theta(z)$ are modelled by deep neural networks.

Supposing that the model approximating the decoder function $\mu_\theta(z)$ is sufficiently expressive, the shape of the prior distribution $p(z)$ does not really matter, and it is traditionally assumed to be a normal distribution $p(z) = G(0,I)$.

Under these assumptions, the term $D_{KL}(q_\phi(z|x)||p(z))$ is the KL-divergence between two Gaussian distributions $G(\mu_\phi(x),\sigma^2_\phi(x))$ and $G(0, I)$ that has the following closed form expression:
\begin{equation}\label{eq:closed-form}
\begin{array}{l}
    D_{KL}(G(\mu_\phi(x),\sigma_\phi(x)),G(0,I)) = \\
    \hspace{1cm}\frac{1}{2} \sum_{i=1}^k \mu_\phi(x)^2_i + \sigma^2_\phi(x)_i-log(\sigma^2_\phi(x)_i) -1
\end{array}
\end{equation}
where $k$ is the dimension of the latent space. 

Coming to the reconstruction loss $\EX_{q_\phi(z|x)} [ \log p_\theta(x|z)]$, under the Gaussian assumption, the logarithm of $p_\theta(x|z)$ is proportional to the quadratic distance between $x$ and its reconstruction $\mu_\theta(z)$; the variance of this
Gaussian distribution can be understood as a parameter balancing
the relative importance between reconstruction error and KL-divergence
\cite{tutorial-VAE,balancing}.

\begin{figure}[ht]
\begin{center}
\includegraphics[width=.4\textwidth]{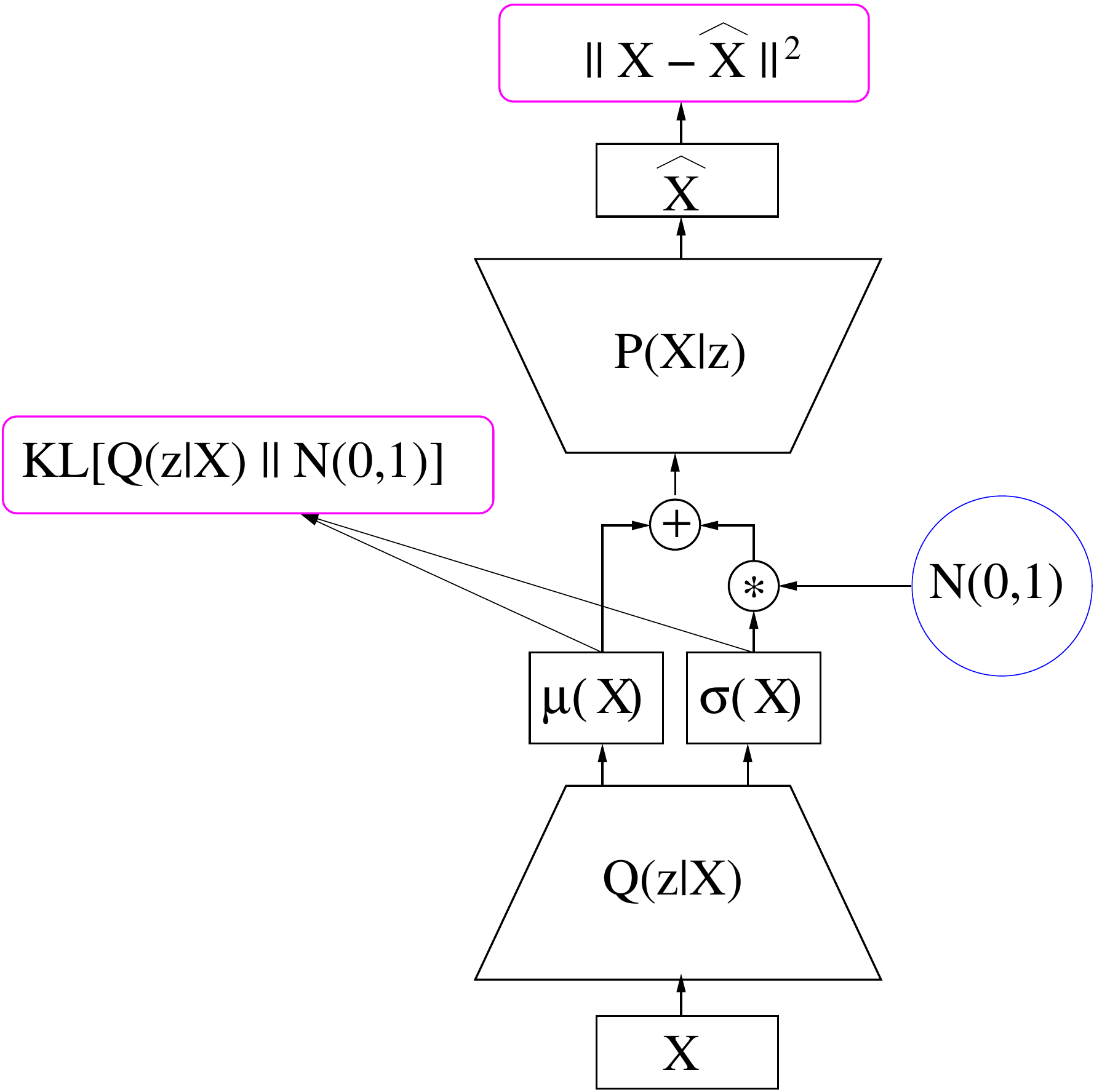}
\caption{VAE architecture}
\label{fig:vae}
\end{center}
\end{figure}

The problem of integrating sampling 
with backpropagation during training is addressed by the so called {\em reparametrization trick} \cite{Kingma13,RezendeMW14}: 
sampling is performed using 
a standard distribution outside of the backpropagation flow and
this value is rescaled with $\mu_\phi(x)$ and $\sigma_\phi(x)$.

It is important to stress that sampling at training time
has no relation with ancestral sampling for generation
(we do not have $x$ at generation time!).
The purpose of sampling at training time is to provide estimates for the main moment of $q(z|x)$, that are then 
subject to KL-regularization. In turn, the final goal of
this regularization is to bring the marginal inference distribution $q(z) = \EX_{x \in\D}q(z|x)$ close to the prior $p(z)$.

\subsection{The dimension of the latent space}
\label{sec:sparsity}

A critical aspect of VAEs is the dimension of the latent space. Typically, having many latent variables reduces the compression loss and improves reconstruction. 
However, this may not result in an improvement of the
generative model: more variables we have and the harder is to ensure their independence and force them to assume the desired prior distribution. 
We may try to
tame them by strengthening the KL-regularization component in the loss function, in the spirit of a $\beta$-VAE \cite{beta-vae17,understanding-beta-vae18}, but this typically results in the {\em collapse} of the less 
informative variables, that get completely ignored by the
decoder \cite{BurdaGS15,overpruning17,Trippe18,PosteriorCollapse,sparsity}. A collapsed variable $z$ has a very characteristic
behaviour: since it is ignored by the decoder, it is
free to minimize KL-regularization, with a mean value 
$\mu_z(x) = 0$ and a variance $\sigma^2_z(x)=1$ for any 
$x$. 

A more expressive architecture may typically result in a 
better exploitation of latent variables, allowing to 
work with a larger number of them. The dimension of the latent space also reflects the complexity of the data manifold: for instance, with non-hierarchical
architectures, it is customary to work with 16 variables
for Mnist, 128 variables for Cifar10 and 64 variables for
CelebA. As we shall see, with a SVAE we can sensibly enlarge
these numbers.

\section{Split-VAE}\label{sec:SVAE}
The general notion of VAE does not impose any 
specification on the architecture of the encoder
and the decoder, and many different variants have been
investigated in the literature: dense, convolutional, with residuality, with autoregressive flows, hierarchical, and so on (see e.g. \cite{variationsVAE,VAEGreen} for a discussion).

A Split-VAE (SVAE) is just another architectural variant: we do not touch the theory or the loss function.
In a SVAE the output $\hat{x}$ is computed as a weighted sum 
\[\hat{x}=\sigma \odot \hat{x_1} + (1-\sigma) \odot \hat{x_2}\] 
of two generated images $\hat{x_1},\hat{x_2}$, and
$\sigma$ is a {\em learned} compositional map. 
Typically, to turn a vanilla VAE into a Split-VAE it is enough to
change the number of channels of the last layer of the network: 
in case of a grayscale image, passing from 1 to 3, and in case of a color
image passing from 3 to 1+3+3=7 (the $\sigma$ map and two color images $\hat{x_1},\hat{x_2}$).
The compound image $\hat{x}$ can be computed internally of externally to
the network, as part of the loss function. 
Since the increased number of channels only concerns the very last layer, the total number of parameters remains comparable to the vanilla version. 

The philosophy underlying a SVAE has been already discussed in the introduction: we merely create an opportunity to be exploited by the model. From a more practical 
perspective, it can be be understood as a way to induce
{\em diversification} in the features learned by the network, via a simple but highly effective self-attention mechanism \cite{attention}.
From this point of view, it is not too far from techniques like
squeeze and excitation \cite{SE} or feature-wise linear modulation \cite{Film}; the main difference is that we
operate on the visible level, and along spatial dimensions. 
This allows us, among other things, to provide intelligible
visualizations of the splitting learned by the network.

One of main characteristic of Split-VAEs is that they allow to work with a sensibly larger number of latent variables: 
32 for Mnist, 200 for Cifar10 and 150 for CelebA. 
This testifies the diversification of latent features, and partially explains the improved generative quality.

\subsection{Encoder-decoder architecture}\label{sec:resnet}
\begin{figure}[ht]
\begin{center}
\begin{tabular}{c}
\includegraphics[width=.33\textwidth]{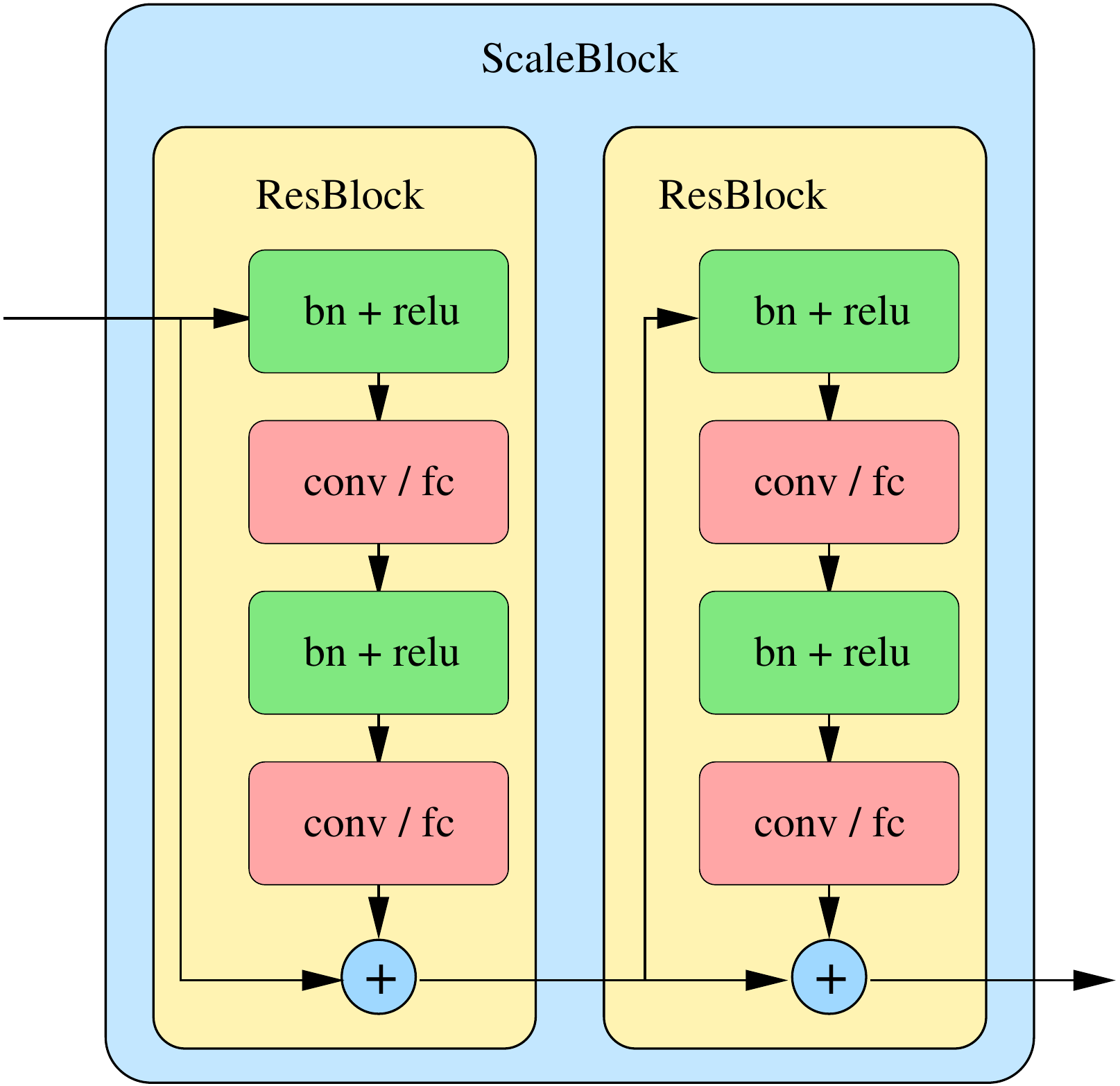} \\
(a) \\
\begin{tabular}{cc}
\includegraphics[width=.16\textwidth]{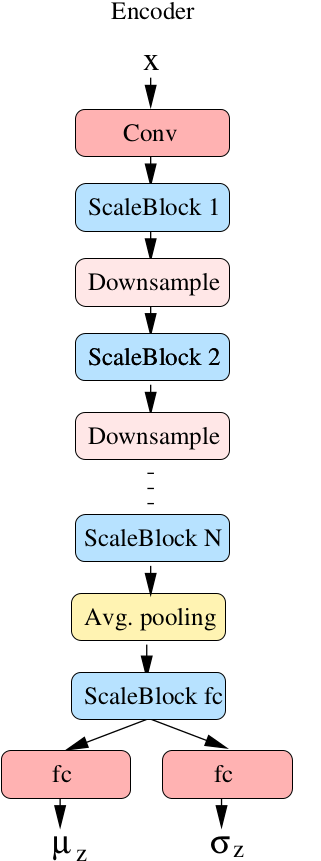} &
\includegraphics[width=.15\textwidth]{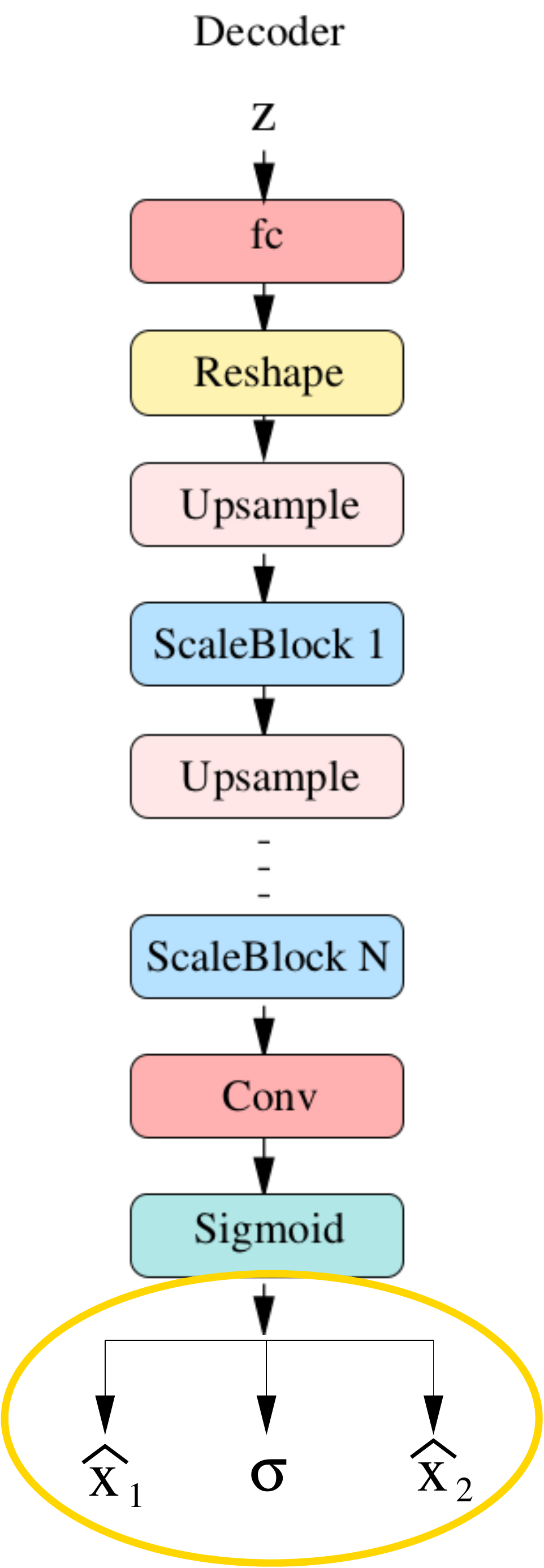}\\
(b) & (c)
\end{tabular}
\end{tabular}
\caption{(a) Scale Block: a Scale Block is a sequence of Residual Blocks intertwined with residual connections. A Residual Block alternates BatchNormalization layers, non-linear units and convolutions. (b) Encoder: the input is progressively downsampled via convolutions, preceded by Scale Blocks. At the final scale, a global average pooling layer extract features that are further processed via dense layers to compute mean and variance for latent variables. (c) Decoder: the decoder is essentially symmetric. A SVAE only differs in the final layer (circled in the picture): instead of directly producing $\hat{x}$, it produces two images $\hat{x_1}$ and $\hat{x_2}$ and a compositional map $\sigma$, defining $\hat{x}=\sigma\odot\hat{x_1}+(1-\sigma)\odot\hat{x_2}$.}
\label{fig:Resnet}
\end{center}
\end{figure}

For the implementation of the encoder and the decoder we adopted a 
ResNet-like architecture derived from \cite{TwoStage} that we already used in previous works \cite{balancing,VAEGreen}; this allows us to do a fair comparison of the split-technique with previous approaches, without additional biases. 
The network architecture is schematically described in Figure \ref{fig:Resnet}.

The encoder is a fully convolutional model where the input is progressively downsampled for a configurable number of times, 
jointly doubling the number of channels.
Before downsampling, the input is processed by a so called {\em Scale Block},
that is just a sequence of {\em Residual Blocks}. A Residual Block is an alternated sequence of BatchNormalization and spatial preserving Convolutional layers, intertwined with residual connections. 
The number of Scale Blocks at each scale of the image pyramid,
the number of Residual Blocks inside each Scale Block, and the
number of convolutions inside each Residual Block are 
user configurable hyperparameters.

In the encoder, after the last Scale Block, a global
average level extracts spatial agnostic features. These are
first passed through a so called {\em Dense Block} (similar to
a Residual Block but with dense layers instead of convolutions),
and finally used to synthesize mean and variance for latent variables.
The decoder first maps the internal encoding $z$ to a small
map of dimension $4\times 4 \times base\_dim$ via a dense layer
suitably reshaped. This is then up-sampled to the final expected
dimension, inserting a configurable number of Scale Blocks at each scale.

\section{Experimental setting}\label{sec:setting}

We compare the performance of SVAE with state-of-the-art variational autoencoders 
comprising Two-stage models \cite{TwoStage,balancing,VAEGreen}, and 
Regularized Autoencoders \cite{deterministic}. We are not considering models 
relying on normalizing flows, such as \cite{glow,NVAE,morrow2020variational}:
these models typically require {\em thousands} of latent variables making them of relatively little interest from the point of view of representation learning. 

For the comparison, we used traditional datasets, such as MNIST, CIFAR-10 \cite{cifar10}
and CelebA \cite{celeba}; the metrics adopted is the usual 
Frech\`et Inception Distance \cite{FID} (FID), shortly discussed in section~\ref{sec:FID}. All comparative results reported 
Section~\ref{sec:tables} are borrowed from the original publications.

In addition to the FID-score for generated images (GEN field, in Tables), we also 
provide an ex-post estimation of the probability distribution of the latent-space.
This is done through a second VAE in \cite{TwoStage,balancing}, and by fitting
a Gaussian Mixture Model (GMM) in \cite{deterministic} (a normalizing autoregressive flow can be used with a similar purpose \cite{autoregressive16,morrow2020variational}).
Although possibly less expressive, the GMM technique is simple and effective, 
so we use it in our experiments (this aspect is somewhat orthogonal to the content of
this article). The FID-score after resampling in the latent space is reported in the 
GMM entry, in the following Tables.

For each architecture, we also provide the number of parameters as an indicative 
measure of its complexity and energetic footprint (according to recent 
investigations \cite{alpha-flops},
the number of parameters seem to provide a more reliable measure of efficiency than
the number of Floating Point Operations).



\subsection{Frech\`et Inception Distance}\label{sec:FID}
The Frech\`et Inception Distance \cite{FID} (FID) 
does not try to assess the ``quality'' of a single generated sample but 
merely compares the overall probability {\em distribution} of
generated vs. real images. The dimension of the visible space is typically too 
large to allow a direct comparison; 
the main idea behind FID is to use, instead of raw data,
their internal representations generated by some third party, agnostic network. In the
case of FID, the Inception v3 network \cite{InceptionV3} trained on Imagenet is used to this purpose.
The activations that are traditionally used
are those relative to the last pooling layer, resulting in a vector of 2048 features.

Let $a_1$ and $a_2$ be the activations relative to real and generated images, and $\mu_i, i=1,2$,
$C_i, i=1,2$ their empirical mean and covariance matrix, respectively. Then the Fr\`echet Distance between $a_1$ and $a_2$ is the just the squared Wasserstein distance, namely:
\begin{equation}\label{eq:FID}
\begin{array}{l}
FID(a_1,a_2) = \\
\hspace{.5cm}= ||\mu_1 - \mu_2||^2 + Tr(C_1 + C_2 - 2(C_1*C_2)^{\frac{1}{2}})
\end{array}
\end{equation}
where $Tr$ is the trace of the matrix.  

\section{Numerical Results}\label{sec:tables}
In this section we give numerical results relative to Mnist, CIFAR-10 and CelebA, discussing the training process and the relevant hyperparameters.

\subsection{Mnist}
In the case of Mnist we worked with a latent space of dimension
32, in contrast with the traditional dimension of 16. We tested 
several versions, with different balancing $\beta$-factor between 
reconstruction and KL-regularization: we report results for
$\beta=8$ and $\beta=3$.
Even when starting with the relatively high balancing factor $\beta=8$, the number of inactive variables at the end of training is low: between 2 and 5.
In all our experiments, the $\beta$-factor is progressively reduced along training in order to preserve the initial
balance between the two components, as described in \cite{balancing}.

In the case of Mnist, both syntactic and semantic decomposition 
(see Figures~\ref{fig:mnist_syntactic} and \ref{fig:mnist_semantic}) usually give good results on the compound image, slightly better 
in the latter case. 

Training lasted 300 epochs, using Adam optimizer with a learning rate of $1.0e-3$. 
Numerical results in terms of FID-scores are given in Table\ref{tab:mnist-results}. 

\begin{table}[h!]

\begin{center}
\begin{tabular}{c|c|c|c}\hline
      model & params & GEN & GMM \\\hline
RAE-GP \cite{deterministic} & 22,386,449 & $22.2 $ & $11.5$  \\\hline
RAE-L2 \cite{deterministic} & 22,386,449 & $22.2 $ & $8.7 $  \\\hline
RAE-SN \cite{deterministic} & 22,386,449 & $19.7 $ & $11.7 $  \\\hline
2S-VAE, \cite{TwoStage} & 9,032,769 & $21.9 $ & $12.6$\\\hline
2S-VAE, \cite{balancing} & 9,032,769 &  $21.8 $ & $11.8$\\\hline
SVAE ($\beta=8$) & 12,793,667 & $20.4$ & $7.3 (20 c.)\; 7.2(100 c.)$\\\hline
SVAE ($\beta=3$) & 12,793,667 & $41.1$ & $7.2 (20 c.)\; \mathbf{4.8} (100 c.)$\\\hline
\end{tabular}
\end{center}
\caption{\label{tab:mnist-results}MNIST: FID scores for generated images (GEN).
The GMM value refers to ex-post estimation of the latent space distribution via a Gaussian Mixture Model in the spirit of \cite{deterministic}. For MNIST, we use a GMM with 20 components.}
\end{table}
In \cite{deterministic}, they observed that 
enlarging the number of components beyond 10 was not beneficial.
However, in our case, presumably due to the larger number of latent variables, we found convenient to use a larger mix.  In Table\ref{tab:mnist-results} we compare the cases of
$20$ and $100$ components. To have an acceptable generative score
before GMM-resampling we need to work with a high $\beta$ factor, 
like e.g. $\beta=8$; however, resampling compensates the
need of regularization, and we obtain the best results with $\beta=3$
and 100 components.

\begin{figure}[h!]
\begin{center}
\includegraphics[width=\columnwidth]{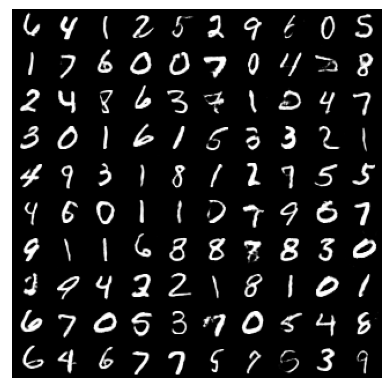}
\caption{Examples of generated images for MNIST}
\label{fig:mnist_square}
\end{center}
\end{figure}

See Figure~\ref{fig:mnist_square} for examples of generated Mnist-like digits.

\begin{figure*}[ht!]
\begin{center}
\includegraphics[width=\textwidth]{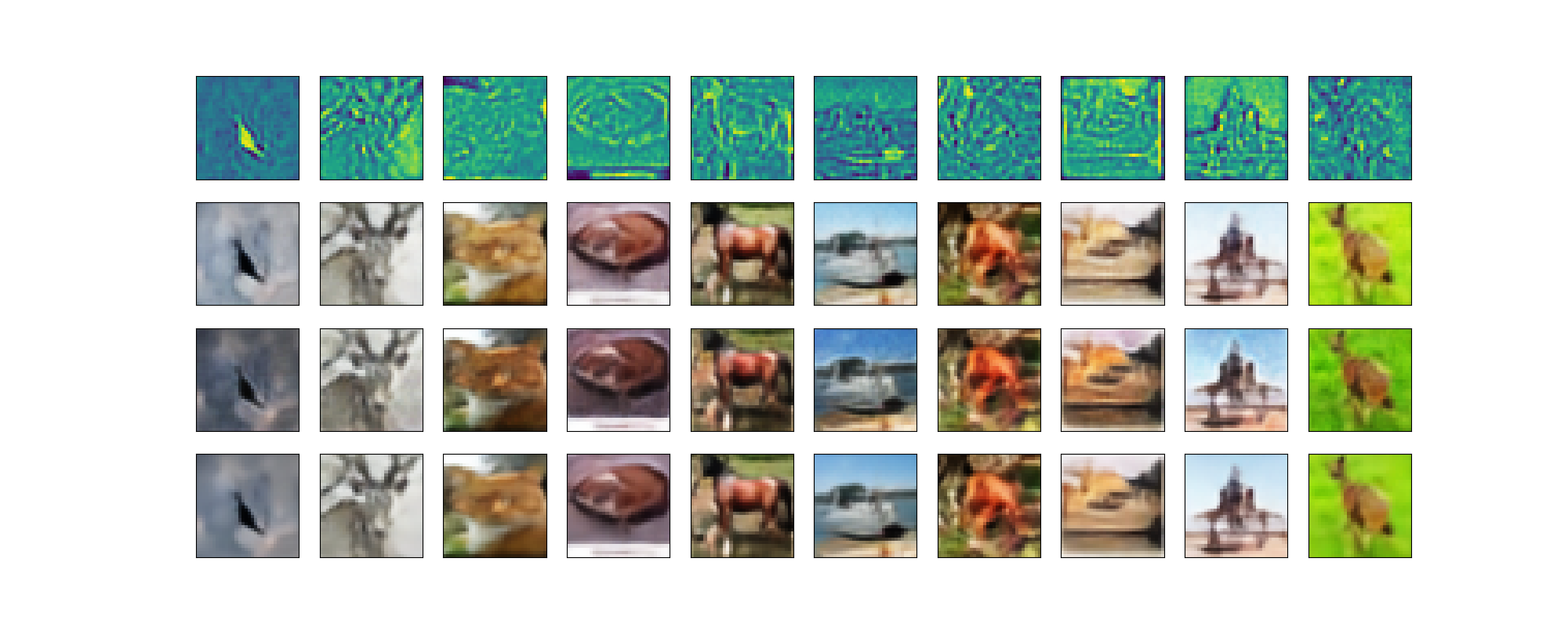}
\caption{Example of reconstructed images in CIFAR-10. The network clearly struggles to derive 
a ``sense'' out of the pictures, but fails. Without the underlying images, contours would be hardly recognizable. Enlarging the resolution does not seem to help.}
\label{fig:cifar_semantic}
\end{center}
\end{figure*}

\subsection{Cifar10}
\begin{figure}[ht!]
\begin{center}
\begin{tabular}{cc}
\includegraphics[width=.45\columnwidth]{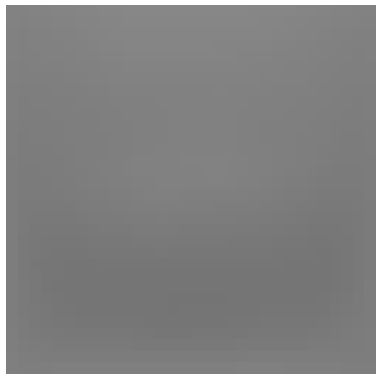} &
\includegraphics[width=.45\columnwidth]{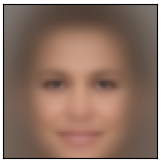} \\
mean CIFAR-10 & mean CelebA
\end{tabular}
\caption{Comparison of the mean image for CIFAR-10 (left) and CelebA (right). Cifar-10 is a sensibly more complex and randomic dataset.}
\label{fig:mean}
\end{center}
\end{figure}

CIFAR-10 confirms its somehow pathological nature. The complexity of the
dataset can be readily appreciated by looking at the pictures in 
Figure \ref{fig:mean} where we compare the mean images for CIFAR-10 and CelebA: the former is 
completely gray, while the latter is a relatively well 
defined \say{average} face (also observe, by the way, the strong bias of the CelebA dataset towards feminine, frontal, 
young, smiling faces).  

Another interesting indicator of the complexity of CIFAR-10 is given by the FID score between each category and the full dataset (we derive 10000 image per category by flipping images in the training set, and compare them with the test-set). The score is extremely high, in spite of the fact that the \say{texture} is apparently quite similar. The FID score \say{magically} drops to 5.3 for a random mix. 

\begin{table}[ht]
\begin{center}
\begin{tabular}{ll|ll|ll|ll}
label & fid & label & fid & label & fid & label & fid \\\hline
plane & 80.7 & car  & 98.3  & bird & 65.0 & cat & 60.8 \\
deer & 66.3 & dog & 74.3 6 & frog & 100.2 & horse & 84.3 \\
ship & 97.7 & truck & 107.5 & \multicolumn{2}{r}{mix} & \multicolumn{2}{l}{5.3} \\\hline
\end{tabular}
\end{center}
\caption{FID score for each one of the CIFAR-10 categories versus the whole dataset. Observe the high values, in comparison with the
extremely low score of the whole dataset vs. itself (mix).}
\label{tab:fid_classes}
\end{table}


Coming to SVAE, in the case of CIFAR-10, splitting produces results of the kind 
described in Figure~\ref{fig:cifar_semantic}. The FID scores for $\hat{x_1}$ and $\hat{x_2}$ are usually lower to
that of $\hat{x}$ testifying that the network is attempting a \say{semantic} decomposition;
however, our networks failed, apart exceptions, to generate recognizable contours. 
In spite of this problem,
numerical results, reported in Table \ref{tab:cifar10-results} are quite good.

\begin{figure}[ht!]
\begin{center}
\includegraphics[width=\columnwidth]{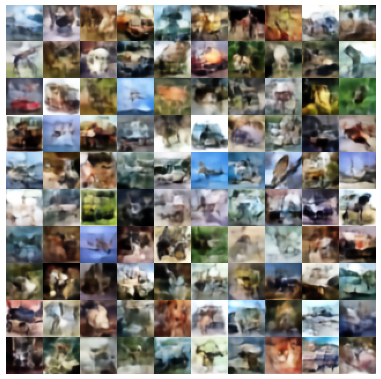}
\caption{Examples of generated images for Cifar-10. 
}
\label{fig:cifar_square}
\end{center}
\end{figure}

\begin{table}[ht!]
\begin{center}
\begin{tabular}{c|c|c|c}\hline
      model & params & GEN & GMM \\\hline
RAE-GP \cite{deterministic} & 30,510,339 & $83.0$ & $74.2$ \\\hline      
RAE-L2 \cite{deterministic} & 30,510,339 & $80.8$ & $74.2$ \\\hline
RAE-SN \cite{deterministic} & 30,510,339 & $84.2$ & $74.2$ \\\hline
2S-VAE \cite{TwoStage}      & 27,766,275 & $76.7$ & $72.9$\\\hline
2S-VAE, \cite{balancing}    & 27,766,275 & $80.2$ & $69.8$ \\\hline
SVAE                        & 13,061,143 & $81.9$ & $\mathbf{69.5}$ \\\hline

\end{tabular}
\end{center}
\caption{\label{tab:cifar10-results}CIFAR-10: summary of results.}
\end{table}
These results have been obtained exploiting a latent space of dimension 200 (in contrast with the traditional dimension of 128)
and a balancing factor $\beta=3$ between reconstruction and KL-regularization.
Training lasted 110 epochs (fast!), using Adam optimizer with an initial learning rate 
of $1.0e-3$. 
For ex-post re-estimation of the distribution of the latent space we used a GMM with 100 components. Examples of generated CIFAR-10-like images are given in Figure \ref{fig:cifar_square}.

\subsection{CelebA}
\label{sec:celeba}
\begin{figure}[ht!]
\begin{center}
\includegraphics[width=\columnwidth]{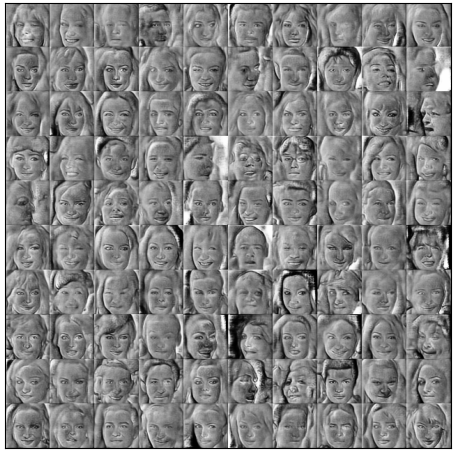}
\caption{Example of {\bf generated} boolean maps. The quality and
precision of contours is both unexpected and remarkable.}
\label{fig:celeba_masks}
\end{center}
\end{figure}

The splitting technique for CelebA works particularly well, automatically producing remarkable ``semantical'' maps similar to drawings (see Figures \ref{fig:celeba_semantic} and \ref{fig:celeba_masks}). The quality and precision in the design of details is impressive and largely unexpected. 

Comparative values are reported in Table \ref{tab:celeba-results}. In this case 
we provide fid scores for $\hat{x}$, $\hat{x_1}$, $\hat{x_2}$.
We worked with a latent space of dimension 150, in contrast with more traditional dimensions like 64 or 128. The initial $\beta$-factor was 3.

\begin{table}[h!]

\begin{center}
\begin{tabular}{c|c|ccc|c}\hline
      model & params & \multicolumn{3}{c|}{GEN} & GMM  \\\hline
RAE-GP \cite{deterministic} & 30,510,339  & \multicolumn{3}{c|}{$116.3$}  & $45.3$ \\\hline
RAE-L2 \cite{deterministic} & 30,510,339  & \multicolumn{3}{c|}{$51.1$} & $47.9$  \\\hline
RAE-SN\cite{deterministic} & 30,510,339  & \multicolumn{3}{c|}{$44.7 $} & $40.9$  \\\hline
2S-VAE, \cite{TwoStage} & 27,766,275 & \multicolumn{3}{c|}{$60.5$} & $44.4$\\\hline
2S-VAE, \cite{balancing} & 27,766,275 &  \multicolumn{3}{c|}{$43.6$} & $38.6$\\\hline
SVAE & 28,989,363 & $54.8$ & $49.0$ & $45.7$ & $\mathbf{35.1}$\\\hline

\end{tabular}
\end{center}
\caption{\label{tab:celeba-results}CelebA: FID scores. In the case of SVAE, the three GEN values respectively refer to $\hat{x}$, $\hat{x_1}$, $\hat{x_2}$. As before, GMM is the FID result after ex-post estimation of the latent space by means of a Gaussian mixture Model (100 components). 
In the case of SVAE the best result is obtained by taking a random mix
of images from $\hat{x_1}$ and $\hat{x_2}$.}
\end{table}

\begin{figure}[h!]
\begin{center}
\includegraphics[width=\columnwidth]{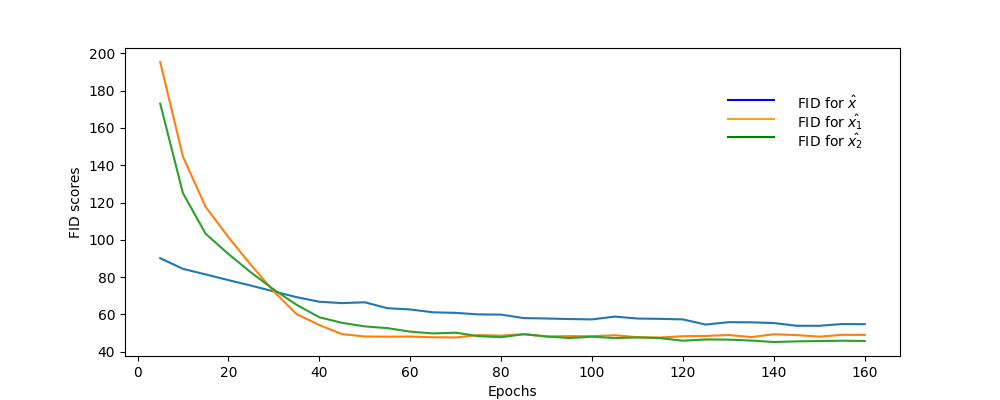}
\caption{Evolution of the FID score for $\hat{x}, \hat{x_1}$
and $\hat{x_2}$ during training.  The score for $\hat{x_1}$ and $\hat{x_2}$ (and for GMM too) is already quite good after 50-60
epochs of training.}
\label{fig:celeba_fid}
\end{center}
\end{figure}

Training lasted 160 epochs, using Adam optimizer with an initial learning rate of $1.0e-3$, 
but already after 40/50 epochs we get excellent FID scores for $\hat{x_1}$ and $\hat{x_2}$.
A typical evolution of FID scores during training is shown in Figure \ref{fig:celeba_fid}; a more thorough investigation is given in the next section.

\begin{figure}[ht!]
\begin{center}
\includegraphics[width=\columnwidth]{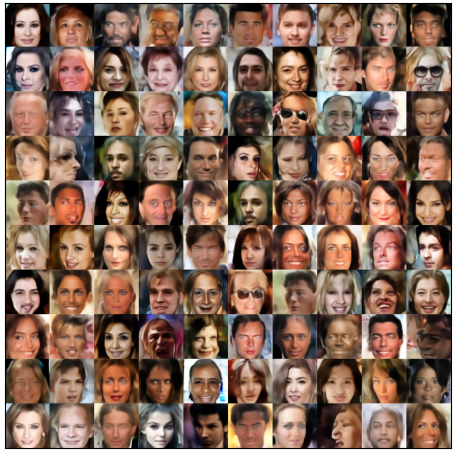}
\caption{Examples of generated images. In the case of CelebA, the quality of samples generated with a variational approach 
should be judged on those details with the
highest variability: hairs, background, accessories. Note also the wide differentiation in pose, illumination, colors, age and expressions.\\}
\label{fig:celeba_square}
\end{center}
\end{figure}

Additional examples of generated CelebA-like images and splitting masks are given in Figures \ref{fig:celeba_square} and \ref{fig:celeba_masks}, respectively.

%

\section{Ablation}
\label{sec:ablation}
The splitting technique is simple and non-invasive. There are however a few additional modifications suggested and induced by splitting - most notably the increased number of latent variables - and one could naturally wonder if this is not the actual source of the observed improvements in FID scores. 

To clarify the point we compared the behaviours of {\em precisely} the same architectures just changing the final layer of the decoder. 

\begin{figure}[h!]
\begin{center}
\includegraphics[width=\columnwidth]{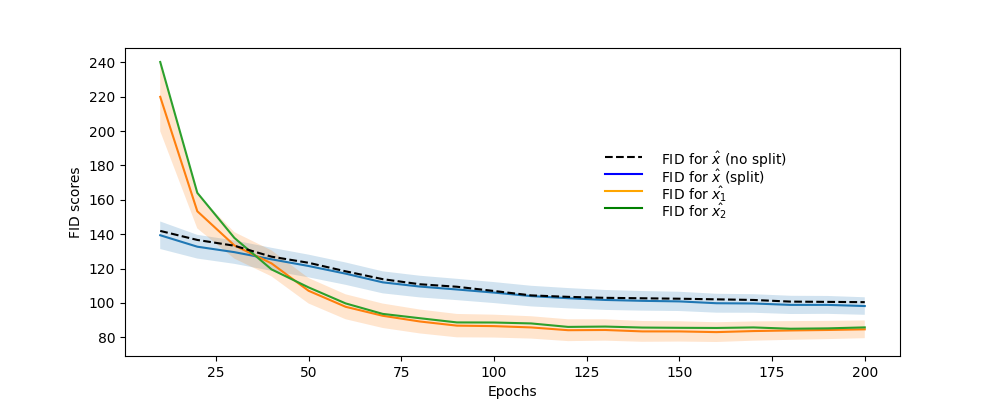}
\caption{CIFAR10: Comparative evolution of the FID score for $\hat{x}$ in vanilla version (dotted line) versus $\hat{x}$
$\hat{x_1}$ and $\hat{x_2}$ for the split-version. 
FID scores have been computed after resampling in the latent space with a GMM with 100 components.
Results
are relative to an average over 5 distinct trainings, along the first 200 epochs. Only
(smoothed) standard deviations for $\hat{x}$ (split version) and
$\hat{x_1}$ are depicted, for readability reasons; the standard deviation for the other FID scores is similar.}
\label{fig:abla_cifar}
\end{center}
\end{figure}

\begin{figure}[h!]
\begin{center}
\includegraphics[width=\columnwidth]{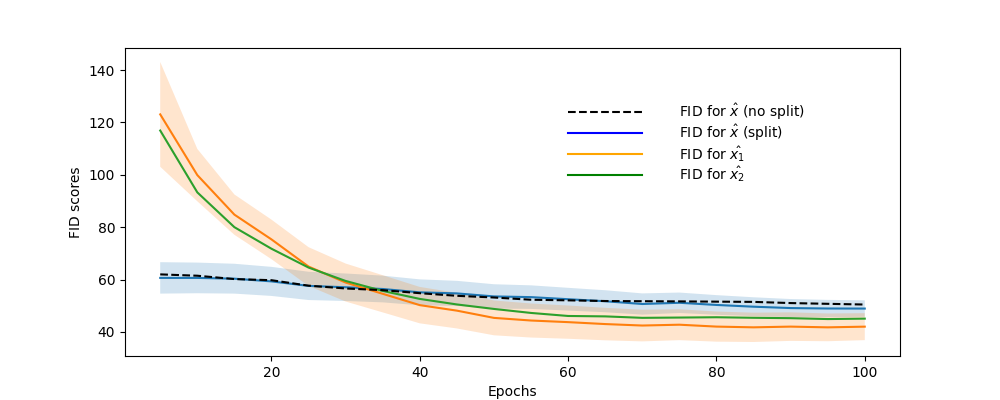}
\caption{CelebA: Comparative evolution of the FID score for $\hat{x}$ in vanilla version (dotted line) versus $\hat{x}$
$\hat{x_1}$ and $\hat{x_2}$ for the split-version. FID scores have been computed after resampling in the latent space with a GMM with 100 components. Results
are relative to an average over 5 distinct trainings, along the first 100 epochs. Similarly to Figure~\ref{fig:abla_cifar}, 
only standard deviations for $\hat{x}$ (split version) and
$\hat{x_1}$ are shown.}
\label{fig:abla_celeba}
\end{center}
\end{figure}

We focused the attention on the most interesting cases of 
CIFAR10 and CelebA, comparing the evolution of the fid score
for $\hat{x}, \hat{x_1}, \hat{x_2}$ for 5 different trainings 
for each dataset. All scores have been computed after resampling in the latent space according to a GMM with 100 components. Results are given in Figures~\ref{fig:abla_cifar} and \ref{fig:abla_celeba}. The improvement for $\hat{x}$ in the case
of the split-network is marginal, but the Fid score for 
$\hat{x_1}$ and $\hat{x_2}$ is {\em significantly} smaller, and frequently even smaller than the reconstruction FID, that is traditionally supposed to be a lower bound for this metric on
generated samples.

Our understanding of the phenomenon is that splitting allows the network to decompose each image towards higher-density regions in
the given neighbourhood, hence creating more realistic variants 
of the usual ``average" result. 

On the other side, it is also interesting to remark
the high sensibility of the FID score to apparently minor 
modifications of generated images (a phenomenon already pointed
out in \cite{varianceloss}).

\section{Conclusions}\label{sec:conclusions}
In this article, we introduced the notion of Split Variational AutoEncoder (SVAE).
In a SVAE the output $\hat{x}$ is computed as a weighted sum $\sigma \odot \hat{x_1} + (1-\sigma) \odot \hat{x_2}$ where $\hat{x_1}, \hat{x_2}$ are two distinct generated
images, and $\sigma$ is a learned compositional map. A Split VAE is trained as a normal VAE: no additional loss is added over the split images $\hat{x_1}$ and $\hat{x_2}$. 
Splitting is meant to offer to the network a way to generate 
variants of the expected result, in the attempt of overcoming
the averaging problem inherent to the adoption of a loglikelihood
loss function. At the same time, the network may specialize 
its generative capabilities towards more oriented and specific 
subsets of the data manifold, possibly learning additional and differentiated features. As a side result, even with a relatively
high balancing factor for KL-regularization, the variable collapse phenomenon is less constraining, and the possibility of exploiting a larger number of latent variables improve the quality and diversity 
of generated samples. 
This has been experimentally confirmed on traditional benchmarks such as Mnist, Cifar10 and CelebA. The SVAE architecture systematically improves over its vanilla counterpart,
and outperforms state-of-the-art loglikelihood-based generative models such as Two-Stage architectures or Regularized autoencoders. We intentionally avoided to test the architecture
on high-resolution datasets such as CelebA-HQ \cite{InvidiaGAN18}, mostly for ethical and ecological reasons: they are too demanding in terms of computational resources. 
We think that there 
are a lot of interesting problems to be investigated and solved even on relatively cheap
datasets, so there is no actual need to move to high-resolution domains. 

As for future developments of this work, a particularly interesting research direction 
seems to be the possibility to add {\em control}
over the splitting operation, possibly segmenting the input image in other interesting 
and meaningful components, and specializing subnets for their respective processing.

\bigskip
\noindent
{\bf Code}
The code relative to this work is available on Github in the following repository: \href{https://github.com/asperti/Split-VAE}{\url{https://github.com/asperti/Split-VAE}}
Pretrained weights for the models discussed in the article
are available at the following page: \href{https://www.cs.unibo.it/~asperti/SVAE.html}{\url{https://www.cs.unibo.it/~asperti/SVAE.html}}.

\bigskip
\noindent
{\bf Conflict of Interest}
On behalf of all authors, the corresponding author states that there is no conflict of interest.

\bibliographystyle{plain}
\bibliography{variational}


\begin{IEEEbiography}[{\includegraphics[width=1in,height=1.25in,clip,keepaspectratio]{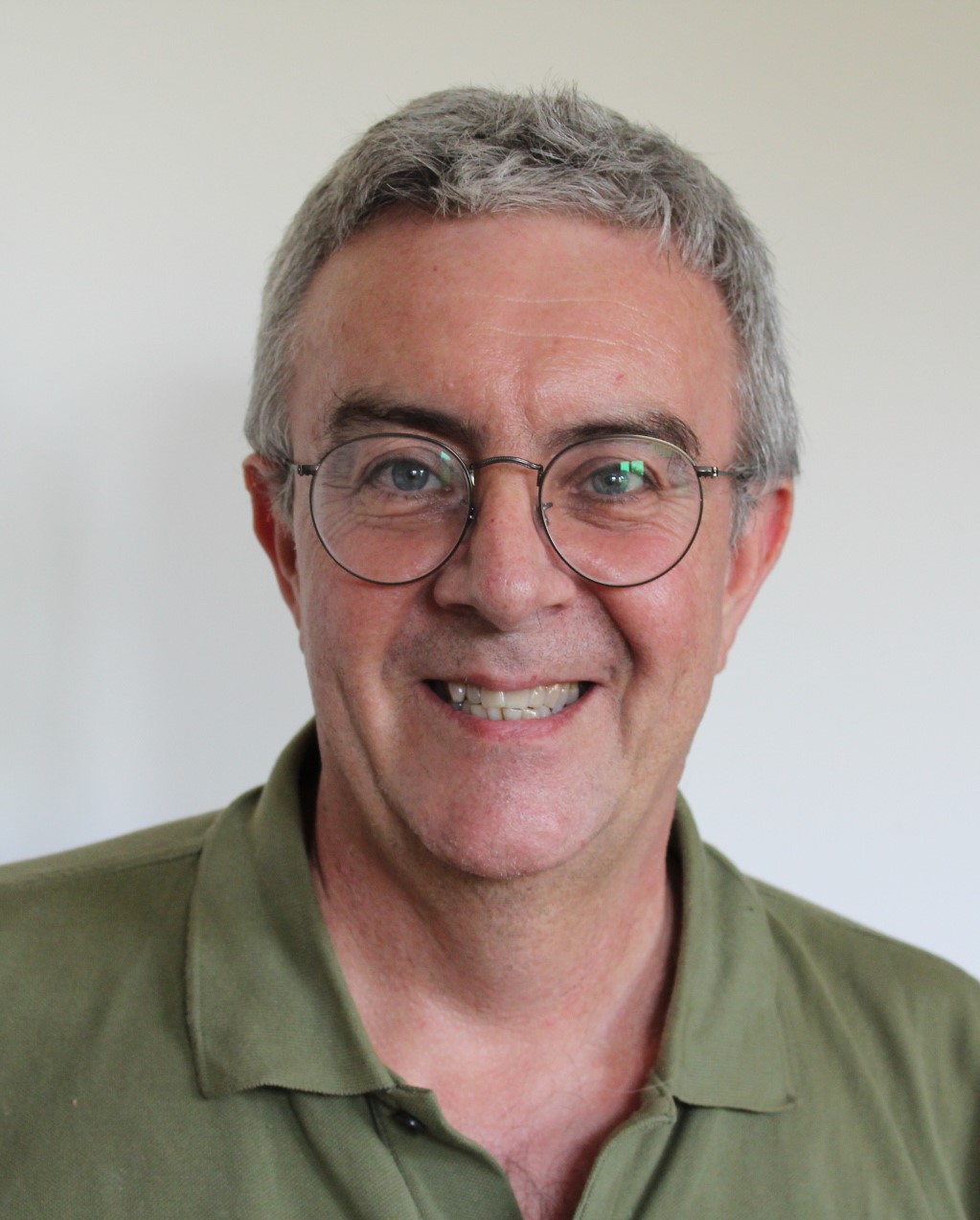}}]{Andrea Asperti} was born in Bergamo in 1961. He received the Ph.D in Computer Science from the University of Pisa in 1989. He is currently Full Professor at the University of Bologna, where he teaches courses of Machine Learning and Deep Learning.

He has been Head of the Department of Computer Science from 2005 until 2007. In the period 2000-2007, he acted as a member of the Advisory Committee of the World Wide Web Consortium. He was responsible for several national and international projects.

He is author of three books, and a number of scientific publications in international peer reviewed conferences and journals. 

His most recent research interests focus on Deep Learning, Generative modeling and Deep Reinforcement Learning.
\end{IEEEbiography}
\begin{IEEEbiography}[{\includegraphics[width=1in,height=1.25in,clip,keepaspectratio]{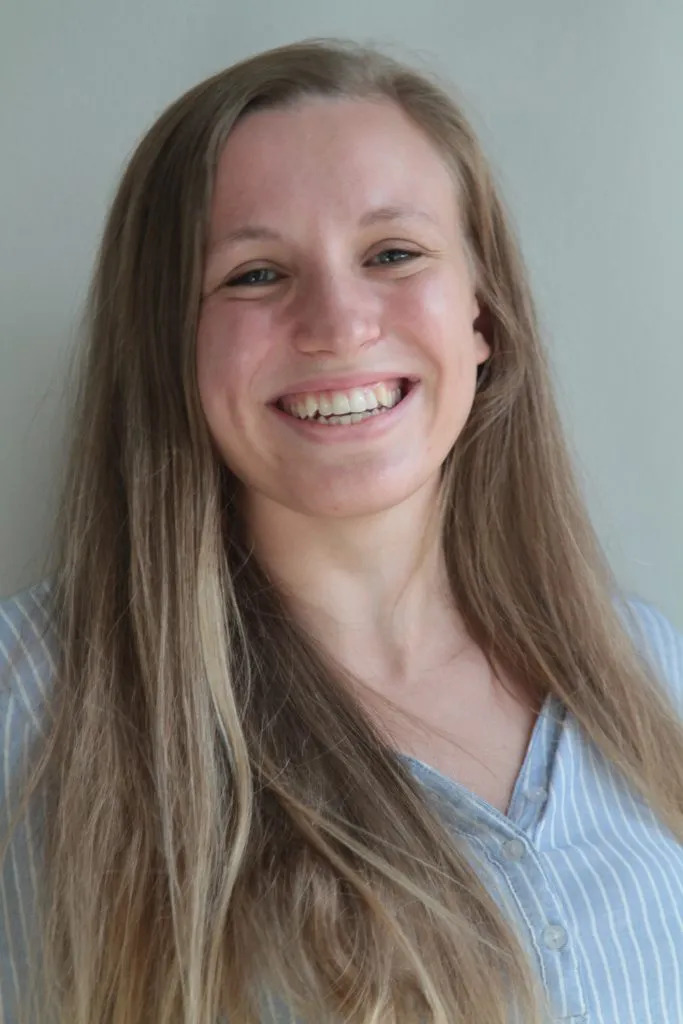}}]{Laura Bugo} 
Laura was born in Bologna, Italy, in 1995. She received the B.S. degree in computer science from the University of Bologna in 2019. She is currently pursuing the master’s degree at the University of Bologna. Her research interests include Artificial Intelligence applied to improvement of well-being for most vulnerable people. She is also an athlete of judo kata, she has won a bronze medal at European Kata Championships in 2021, 3 silver medals at European Kata Championships U24 and a silver medal at World Judo Kata Grand Slam U35.
\end{IEEEbiography}
\begin{IEEEbiography}[{\includegraphics[width=1in,height=1.25in,clip,keepaspectratio]{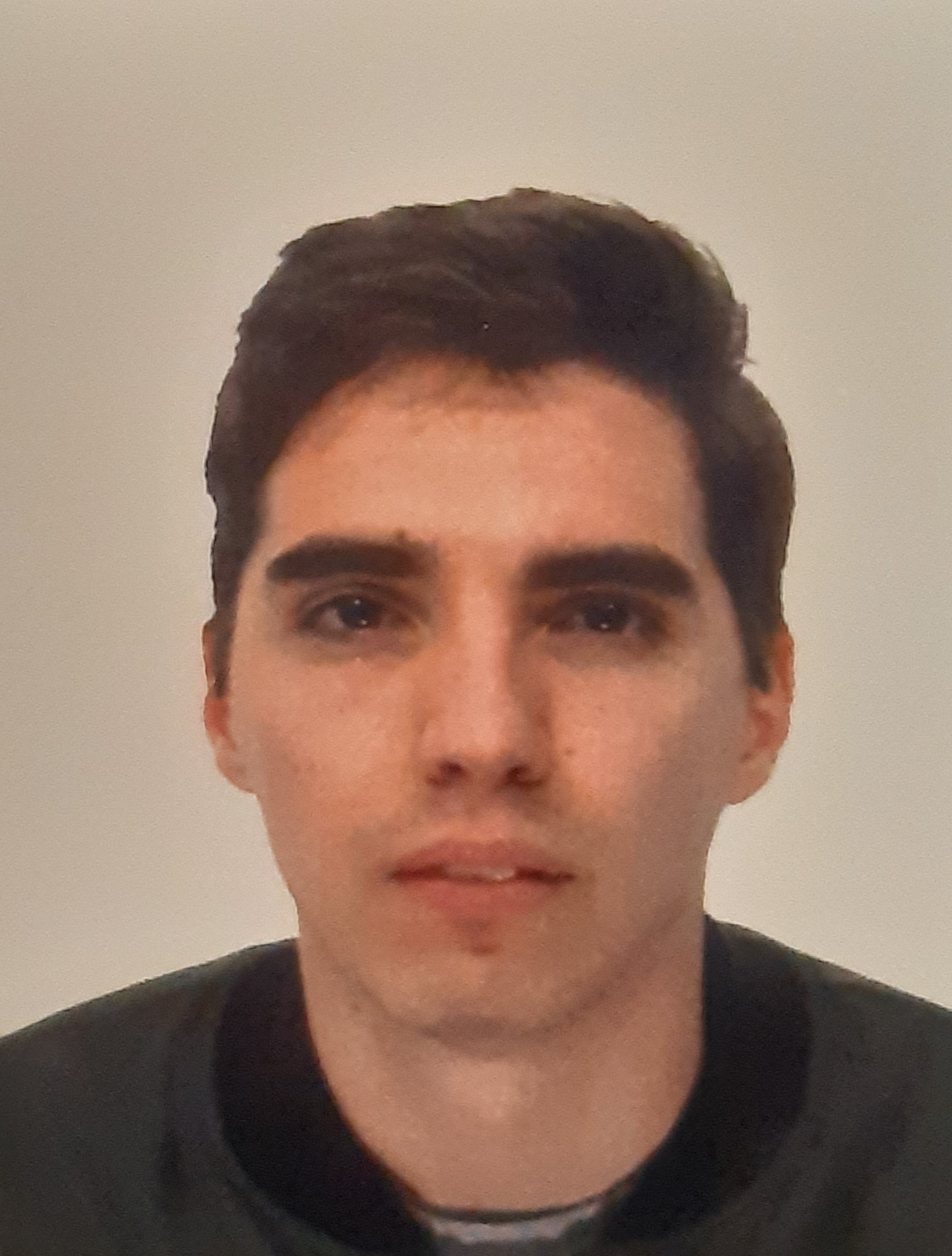}}]{Daniele Filippini} 
Daniele Filippini  was born in Ostiglia , Italy, in 1996 . He received the B.S. degree in information science for management from the University of Bologna in 2019. He is currently pursuing the master’s degree in computer science at University of Bologna. His research interests include artificial intelligence and deep learning.
\end{IEEEbiography}

\EOD 

\end{document}